# A deep artificial neural network based model for underlying cause of death prediction from death certificates

## Authors

Louis Falissard (corresponding author):

CépiDc Inserm, Paris Saclay University, le Kremlin Bicêtre, France
Postal address: 31 rue du Général Leclerc, 94270, Le Kremlin Bicêtre, France
Email address: louis.falissard@inserm.fr
Phone number: +33679649178

Claire Morgand: CépiDc, Inserm, Le Kremlin Bicêtre, France

Sylvie Roussel : CépiDc, Inserm, Le Kremlin Bicêtre, France

Claire Imbaud: CépiDc, Inserm, Le Kremlin Bicêtre, France

Walid Ghosn: CépiDc, Inserm, Le Kremlin Bicêtre, France

Karim Bounebache: CépiDc, Inserm, Le Kremlin Bicêtre, France

Grégoire Rey: CépiDc, Inserm, Le Kremlin Bicêtre, France

# 1 Introduction

The availability of up-to-date, reliable data on causes of death is a matter of significant importance in public health related disciplines. As an example, the monitoring of leading causes of deaths is an important tool for public health practitioner and has a considerable impact on health policy related decision making processes[1–6]. The collection of said data, however, is a complex, time consuming process that usually involves the coordination of many different actors, starting from medical practitioners writing death certificates, to the final statistics' diffusion by public institutions. One example of non-trivial task involved in this process is the identification of the underlying cause of death from the chain of event reported by the medical practitioner in the death certificate[7]. According to the International Statistical Classification of Diseases and Related Health Problems, the underlying cause of death is defined as "(a) the disease or injury which initiated the train of morbid events leading directly to death, or (b) the circumstances of the accident or violence which produced the fatal injury" [8]. Since the underlying cause of death is adopted as the cause for tabulation of mortality statistics, extracting it from death certificates is of paramount importance.

Nowadays, in order to preserve spatial and temporal comparability, the underlying cause of death is usually identified from an expert system[9] (such as the Iris software[10]), a form of artificial intelligence that encodes a series of WHO-defined coding rules. Unfortunately, these decision systems fail to handle a significant amount of more complex death scenarios, typically including multiple morbidities or disease interactions. These cases then require human evaluation, consequently leading to a time consuming coding process potentially subject to distributional shift across both countries and years, sensibly impairing the statistics' comparability.

In the past few years, the field of artificial intelligence has been subject to a significant expansion, mostly led by the recent successes encountered in the application of deep artificial neural networks based predictive models in various tasks such as, for instance, image analysis, voice analysis or natural language processing. These methods have been known to outperform expert systems, but usually require vast amounts of data on which to train to do so, which is oftentimes prohibitive. On the other hand, a number of countries, including France, have been storing their death certificates, along with their derived underlying causes, in massive databases, thus providing with an optimal setting to use deep learning methods.

The following article formulates the process of extracting the underlying cause of death from death certificate as a statistical predictive modelling problem, and proposes to solve it with a deep artificial neural network. The following section focuses on describing the structured information contained in a death certificate. Section 3 introduces the neural network architecture used for the task of predicting the underlying cause of death. Section 4 reports the results obtained from training the neural network on French death certificate from the years 2000 to 2015 (about 8 million training examples) as well as a comparison with prediction performances obtained using the Iris software, current state of the art for this predictive task and solution used in numerous countries for underlying cause of death coding. Finally, section 5 shows an application of the derived model on opioid overdose related deaths in France.

## 2 Material and Method

### 2.1 Dataset

The dataset used during this study consists of every available death certificate found in the CépiDc database for the years 2000 to 2015, and their associated cause of death, representing over 8 million training examples. These documents record various information about their subjects, with varying predictive power with regard to the underlying cause of death. The following article aims to derive a deep neural network based predictive model explaining the underlying cause of death from the information contained within death certificates:

$$P(UCD|DC) = f(DC)$$

With:

- $DC$ the information contained in a French death certificate
- $UCD$ its corresponding underlying cause of death
- $f$ a neural network based predictive function

In order to model the underlying cause of death from these information, the following items were selected as explanatory variables:

- The causal chain of events leading death
- age
- gender
- year of death

### 2.1.1 Causal chain of death

The causal chain of death constitutes the main source of information available on a death certificate in order to devise its corresponding underlying cause of death. It typically sums up the sequence of events that led to the subject's death, starting from immediate causes (such as cardiac arrest) and progressively expanding into the individual's past to the underlying causes of death. The latter being the target of the investigated predictive model, the information contained in the causal chain of death is of paramount importance to decision process leading to the underlying cause of death's establishment. In order to enforce death statistics comparability across countries, the coding of the underlying cause of death from the causal chain of events is defined from a number of WHO issued rules oftentimes reaching casuistry on more complex situations[11].

The WHO provides countries with a standardized causal chain of events format, which France, alongside every country using the Iris software, follows. This WHO standard asks of the medical practitioner in charge of reporting the events leading to the subject's passing to fill out a two part form in natural language. The first part is comprised of 4 lines, in which the practitioner is asked to report the chain of events, from immediate to underlying cause, in inverse causal order (immediate causes are reported on the first lines, and underlying causes on the last lines). Although 4 lines are available for reporting, they need not all be filled. In fact, the last available lines are rarely used by the practitioner (line 4 was used less than 20% of the time in the investigated dataset).The second part is comprised of two lines in which the practitioner is asked to report any "other significant conditions contributing to death but not related to the disease or condition causing it"[12] that the subject may have been suffering from. Although this part might seem at first sight to have close to no impact on the underlying cause of death, some coding rules ask that the latter should be taken from this part of the death certificate. As an example, the underlying cause of death of an individual with AIDS who died from Kaposi's sarcoma should be coded as AIDS, although this condition acts as a comorbidity and should only appear on the second part of the death certificate. Consequently, this part of the death certificate also presents some vital information for the investigated predictive model, and as such should be included as input variable.

| type | line | text |
|---|---|---|
| | | Sample Certificate 1 |
| Raw causes | 1 | CARDIAC ARREST |
| | 2 | ACUTE CORONARY SYNDROME |
| | 3 | ACUTE OR CHRONIC KIDNEY DISEASE |
| | 4 | DIABETIC NEUROPATHY |
| | 6 | PERIPHERAL ARTERIAL DISEASE; DM II |

Fig.1 Example of causal chain of death as found on a French death certificate. Its corresponding underlying cause of death was defined as "diabetes mellitus type 2, with multiple complications"[13]

In order to counter the language dependent variability of death certificates across countries, a pre-processing step is typically applied to the causal chain of events leading to the individual's death, where each natural language based line on the certificate is converted into a sequence of codes defined by the 10th revision of the International Statistical Classification of Diseases and Related Health Problems (ICD-10). ICD-10 is a medical classification defined by the WHO[8] defining 14199 medical entities[14] (e.g. diseases, signs and symptoms…) distributed over 22 chapters (found in table 2) and encoded with 3 or 4 alpha decimal symbols (one letter and 2 or 3 digits), 7404 of which are present in the investigated dataset. The WHO defined decision rules governing the underlying cause of death process are actually defined from this ICD-10 converted causal chain, and the former is to be reported as a unique ICD-10 code.

The processed causal chain of death, in its encoded format, can be assimilated as a sequence of 6 varying length sequences of ICD-10 codes. In order to simplify both the model and computations, this hierarchical data structure will hereon be assimilated, as seen in figure 2, as a padded 6 by 20 grid of ICD-10 codes, with rows and columns denoting a code's line and rank in line, respectively, 20 being the maximal number of ICD-10 codes found on a causal chain line in all certificates present in the investigated dataset. Several, more subtle approaches to this grid like assimilation were explored prior to the experiment reported in this article, but all yielded models with significantly inferior predictive power. Although this encoding scheme apparently prevents the encoding to handle death certificates with at least one line containing more than 20 codes, the model introduced further itself sees no such limitation. Bigger certificates can be processed without trouble with an appropriately larger code matrix encoding, with theoretically no significant loss in performance, since the model is translation invariant[15].

The question of encoding ICD-10 codes in a statistically exploitable format is another challenge in itself. A straightforward approach would be to factor each ICD-10 code as a 7404 dimensional dummy variable. This simple encoding scheme might however be improved upon, typically by exploiting the ICD-10 hierarchical structure by considering codes as sequences of character. This approach was investigated, but yielded significantly lower results. As a consequence, the results reported in this article only concern the dummy variable encoding scheme.

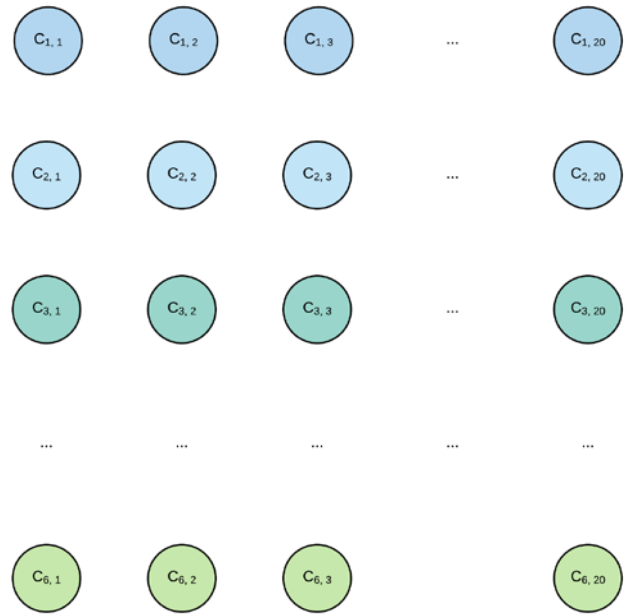

Fig. 2 Causal chain of death encoded as a 3 dimensional tensor. Each node represents an ICD-10 code as a 7404 dimensional dummy variable. Its row and column position respectively denotes the corresponding code's line and rank in the corresponding certificate

### 2.1.2 Miscellaneous variables

From gender to birth town, a death certificate contains various additional information on its subject besides the chain of events leading to death. As some of these items are typically used by both Iris and human coders to decide the underlying cause of death, they present an interest as explanatory variable for the investigated predictive model. After consultation with expert coders, the following items available on French death certificate were selected as additional exogenous variables:

- gender (2 states categorical variables)
- year of death (16 states categorical variables)
- age, factorized into 5 years intervals from subject less than one year old, which were divided into two classes

Strictly speaking, the subject's year of passing should only have a limited effect on the underlying cause of death (aside maybe from years with particularly deadly events, such as the 2003 European heatwave[16]). However, the WHO defined coding rules, as well as their interpretations by human coders slightly evolve over the years. As a consequence, the model should benefit, in term of predictive performance, from being able to differentiate between different years.

## 2.2 Neural architecture

With the death certificate and its selected variables converted into a format enabling analysis, the underlying cause of death extraction task can be solved by estimating its corresponding ICD-10 code's probability density, conditioned on the aforedefined explanatory variables:

$$P(UCD|CCD, A, Y, G, \theta) = f_\theta(CCD, A, Y, G)$$

With:

- $UCD \in \mathbb{R}^{7404}$ the underlying cause of death
- $CCD \in \mathbb{R}^6 \times \mathbb{R}^{20} \times \mathbb{R}^{7404}$ the ICD-10 grid encoded causal chain of death
- $A \in \mathbb{R}^{25}$ the categorized age
- $Y \in \mathbb{R}^{16}$ the year of death
- $G \in \mathbb{R}^2$ the gender
- $f_\theta$ a mapping from the problem's input space to its output space, parameterized in $\theta$ a real-valued vector (typically a neural network)

Although properly defined, the investigated prediction problem still present significant challenges for traditional statistical modelling methods. First, it is expected that the relationship between the input variables and the investigated regressand should be highly non-linear, whereas most statistical modelling techniques are typically used in linear settings. Feed-forward neural networks[17], however, were developed as powerful nonlinear expansions of traditional linear or logistic regressions. The idea behind feedforward neural networks is to fit a linear model to a transformed set of observations where not only the model's parameters, but also the feature transformation, are learnt from the data.

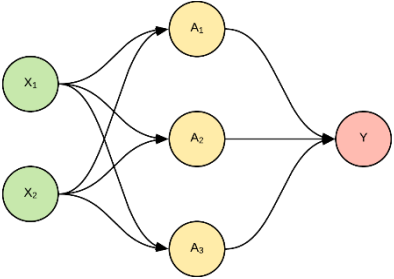

Fig. 3 Example of feed-forward neural network with one hidden layer of 3 neurons

To do so, neural network based methods rely on elementary, differentiably parameterized non-linear functions denoted as neurons, which are typically defined as linear combinations of their inputs injected into a simple, differentiable non-linear function (a logistic function, for instance), as can be seen on figure 3. An artificial neural network can then derive powerful non-linear models by injecting the investigated input variables into a significant amount of neurons (also called neural layer), whose output are then used to fit a traditional linear model such as logistic or linear regression, while adjusting on all tunable parameters, neurons and linear model alike[17], as can be seen in figure 4.

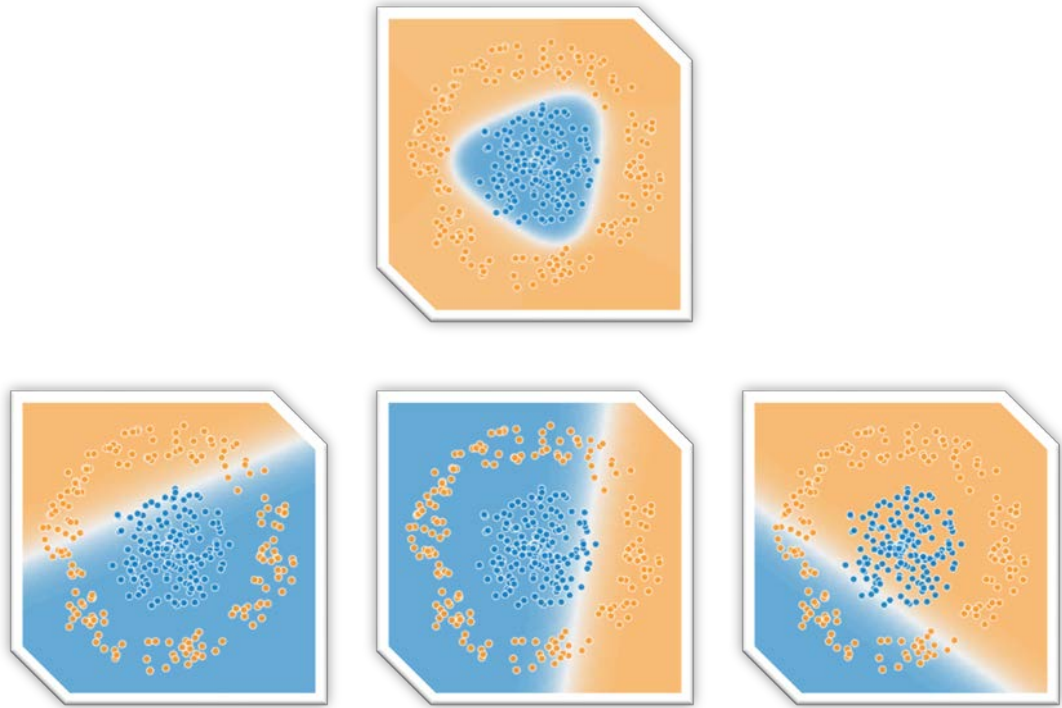

Fig. 4 Neural network as shown in Fig. X applied to a toy example. The top image represents the network's output function. The three bottom images show individual neurons' output. By combining those three carefully chosen linear boundaries, the network is able to model the 'ring-like' dataset it is fed.

This approach can be yet improved upon by stacking several neural layer between the input variable and the final linear model output, progressively increasing the model's number of degrees of freedom and non-linear explanatory power[18] and leading to the creation of a wide family of predictive model, with state of the art performance in a wide variety of tasks, typically in computer vision and natural language processing. Although the currently investigated modelling problem doesn't fall into one of these categories, recent advances in both deeply inspired the neural architecture presented in this article, which can be seen in figure 5 and can be decomposed as follows:

- Linear projections are applied to each one-hot encoded categorical variable[19] (one linear projection is shared for all ICD-10 codes present in the causal chain of death), with all linear projections sharing the same output space dimension

- The miscellaneous variables' projections are added to all of the projected grid's elements

- The resulting grid is used as input to a convolutional neural network[20]

- A multinomial logistic regression (softmax regression) targeting the underlying cause of death is performed on the convolutional neural network's output[21]
- All model parameters (from both the linear projections and the convolutional network) are adjusted by minimizing a cross-entropy objective using gradient based optimization. The model's gradients are computed using the backpropagation method[17].

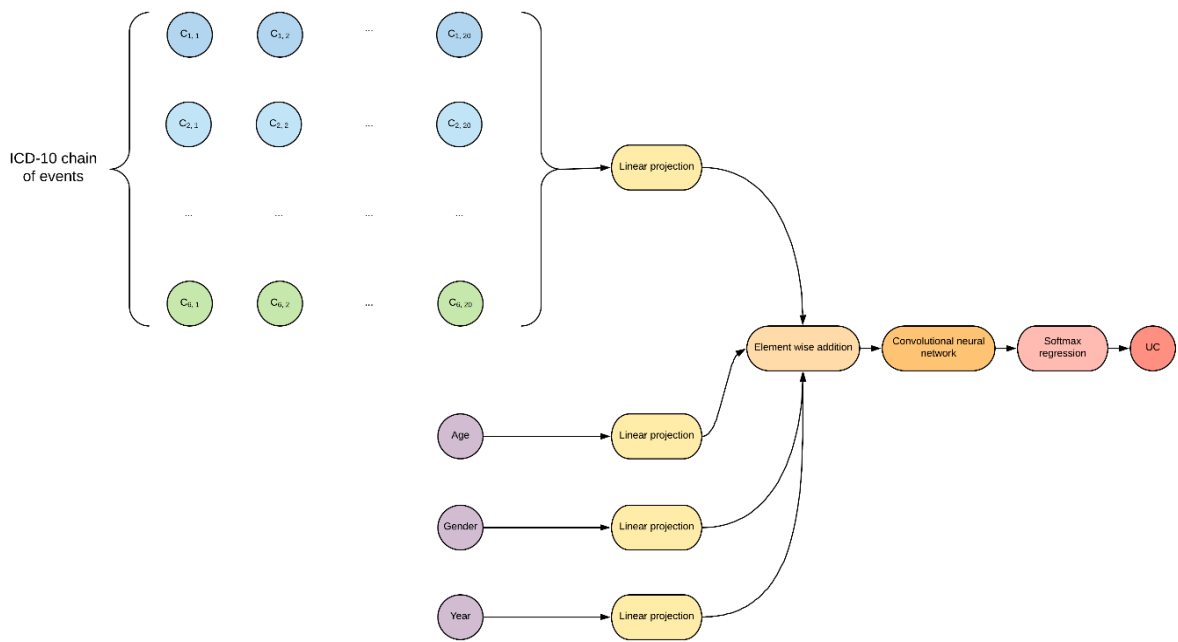

Fig. 5: Overall model architecture

The authors feel that the formal definition of all the model's constituents fall outside the scope of this article. The interested reader will, however, find a complete description of the model in the annex, as well as a fully implemented example (written with Python and Tensorflow) in [22]. We also encourage interested readers to explore the multiple articles that influenced this architecture's design, which are all available in the bibliography[19,23–25].

## 2.3 Training and evaluation methodology

In a similar fashion to what is typically done in generalized linear models, neural network are fit by minimizing a cost function (in this case the cross-entropy function, also used in logistic regression settings, for instance) using gradient based iterative optimization algorithms. However, fitting deep neural models is a process widely known as quite demanding. First, their significant amount of tuneable parameters (up to several hundred millions) makes them quite demanding in term of computational complexity and typically require the use of parallel or distributed computing on both specialized hardware and software. More importantly, the cost functions associated with neural networks have been shown to be sensibly difficult to optimize using classical gradient based methods, and phenomena like vanishing or exploding gradients[26], that have for a long time prevented the successful training of most neural models, still cause complications that should not be under estimated. Consequently, the process of training a deep neural network needs to follow a carefully chosen methodology involving the tuning of several hyper-parameters, usually derived from trial and error.

The investigated model was trained using all French death certificates from years 2000 to 2015. 10,000 certificates were excluded from each year and spread into a validation set for hyper-parameter fine-tuning, and a test dataset for unbiased prediction performance estimation (5000 each), resulting in three datasets with following sample sizes:

- Training dataset: 8553705 records
- Validation and test dataset: 80000 records each

The model was implemented with Tensorflow, a python-based distributed machine learning framework, on two NVidia RTX 2070 GPUs simultaneously using a mirrored distribution strategy. Training was performed using a variant of stochastic gradient descent, the Adam optimization algorithm.

The numerous hyper-parameters involved in the model and optimization process definition were tuned using a random search process. However, due to the significant amount of time required to reach convergence on the different versions of the model trained for the experiment (around 1 week per model) only three models were trained, the results displayed below being reported from the best of them, in term of prediction accuracy on the validation set. The interested reader will find a complete list of the hyper-parameters defining this model in annex. Given the considerably small hyper-parameter exploration performed for the experiment reported in this article, the authors expect that better settings might provide with a slight increase in prediction performance. However, given the successful results obtained and the computational cost of a finer tuning, a decision was taken to not further the exploration.

After training, the model's predictive performance was assessed on the test dataset (excluded prior to training, as mentioned earlier), and compared to the Iris software's, nowadays considered as the state of the art in automated coding and internationally used. In order to ensure a fair comparison between the two systems, Iris' performances were assessed on the test set as well and given the same explanatory variables. As is done traditionally in the machine learning academic literature, the predictive performance is reported in term of prediction accuracy, namely the fraction of correctly predicted codes in the entire test dataset.

The Iris software' automatic coding accuracy was assessed with two distinct values resulting from the software's ability to automatically reject cases considered as too complex to be handled by the decision system. As a consequence, a first accuracy measurement (the lowest one) was assessed considering rejects as ill-predicted cases, while the second one excluded these rejects from the accuracy computation, thus yielding an improved estimate. In order to present the reader with a more comprehensive view of both approaches' performances these accuracy metrics were also derived on a per chapter basis, again on the same test set.

# 3 Results

The neural network based model was trained as aforedescribed for approximately 5 days and 18 hours, and its predictive performance as well as Iris' are reported in table 1.

The neural network based approach to underlying cause of death automated coding significantly outperforms both metrics. Indeed, even when compared to Iris' performance on non-rejected cases, the error rate offered by the proposed approach is 3.4 times lower. This performance difference increases to an eleven fold decrease when including rejected cases in Iris performance.

| Selected approach | Prediction accuracy |
| --- | --- |
| Iris overall accuracy | $0.745\ [0.740, 0.750]$ |
| Iris on non-rejected certificates | $0.925\ [0.921, 0.928]$ |
| Proposed approach | $\mathbf{0.978\ [0.977, 0.979]}$ |

Table 1 Prediction accuracy of Iris and the best derived predictive model, with their corresponding 95% confidence intervals, derived by bootstrap

In addition, Figure 6 shows the model's error rates per ICD-10 chapter, alongside the latter's prevalence. In this plot chapter VII, diseases of the eye and adnexa, appears as a strong outlier in term of error rate. Also not statistically significant (only 3 death certificates among the 80 thousands sampled for the test set have a chapter VII related underlying cause of death), this observation might indicate that the training set does not have a big enough sample size to allow the model to handle extremely rare cases such as chapter VII related death certificates, which might better be handled by a hand crafted, rule based decision system.

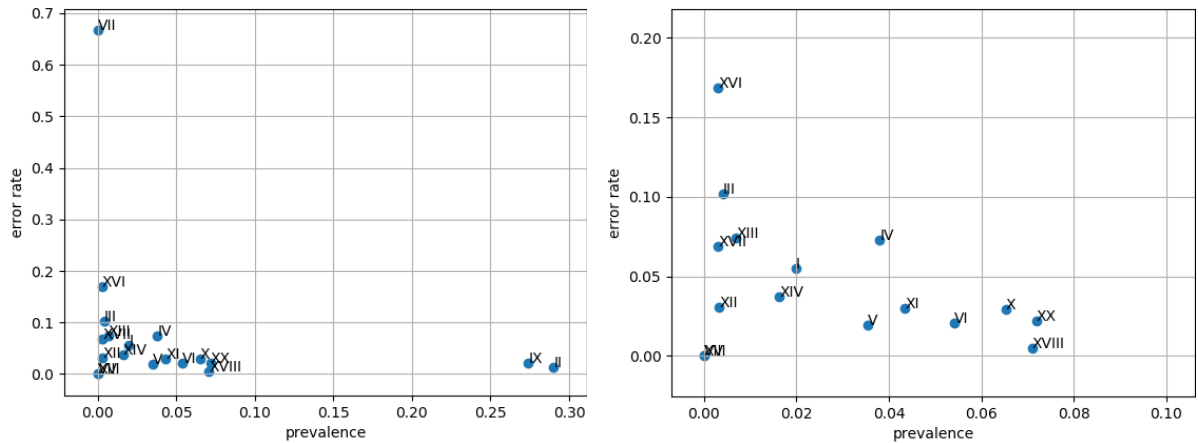

Fig. 6 Left: Prevalence of underlying causes (by ICD-10 chapter) against ICD-10 chapter level model error rate. Right: Zoom on the previous plot's bottom left hand corner

Finally, figure 7 shows the per chapter difference in error rate between the proposed neural network approach and the Iris software (on non-rejected certificates). As previously hypothesized, the Iris software outperforms the deep learning approach on diseases of the eyes and adnexa related death certificates (chapter VII), although still not significantly. Even if the Iris software is beaten in every other chapter, a case should be made from never appearing chapters. Indeed, a number of chapters (namely chapters XIX, XXI and XXII) are not observed as underlying causes in the test dataset, strongly indicating that they might benefit from a set of hand crafted rules as do chapter VII related certificates, if they were to appear in extremely rare cases.

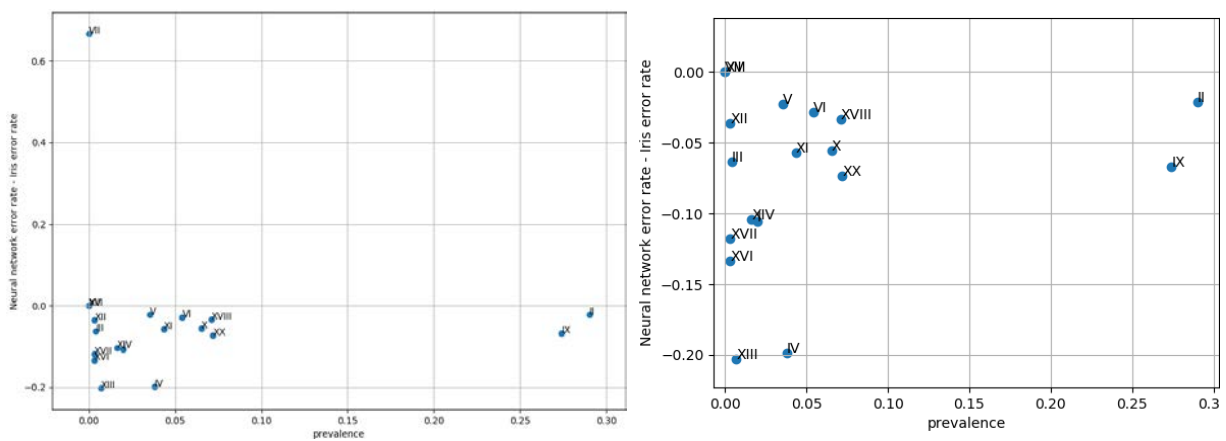

Fig. 7 Left: Difference in error rate between the proposed model and the Iris software versus chapter prevalence as underlying cause. Right: Zoom on the previous plot's bottom left hand corner

## 4   Error Analysis

Although the proposed approach significantly outperforms the current state of the art that is the Iris software, neural network based methods are known to present several drawbacks that can significantly limit their application in some situations. Typically, the current lack of systematic methods to interpret and understand neural network based model and their decision processes can lead the former to perform catastrophically on ill predicted cases, independently from their high predictive performances.

As a consequence, the proposed model behaviour in ill predicted cases require careful analysis. In addition, the system's performance can potentially benefit from such an investigation. For instance, although the model outperforms Iris on average, there might some highly non-linear exceptions that are better fit to rule-based decision systems, in which case a hybrid approach could, by using the best of both world, again yield performance gains.

Although assessing per chapter error rates, as previously shown, constitutes a simple, straightforward approach to understanding the model's weakness, much more can be done to gain insight into the model's behaviour. As an example, it only feels natural, after identifying cases incorrectly predicted by the investigated model, to assess the nature of errors made by the latter. As aforementioned, neural network based classifiers tend to, in misprediction cases, output answer unreasonably far from the ground truth. One should however expect from a good predictive model to, in error cases, output predictions as close as possible to the correct answer. Figure 9 displays an ICD-10 chapter level confusion matrix built from ill predicted test cases, and shows that beside chapter VII, most of the errors remain in the same chapter as the ground truth, indicating some degree of model robustness.

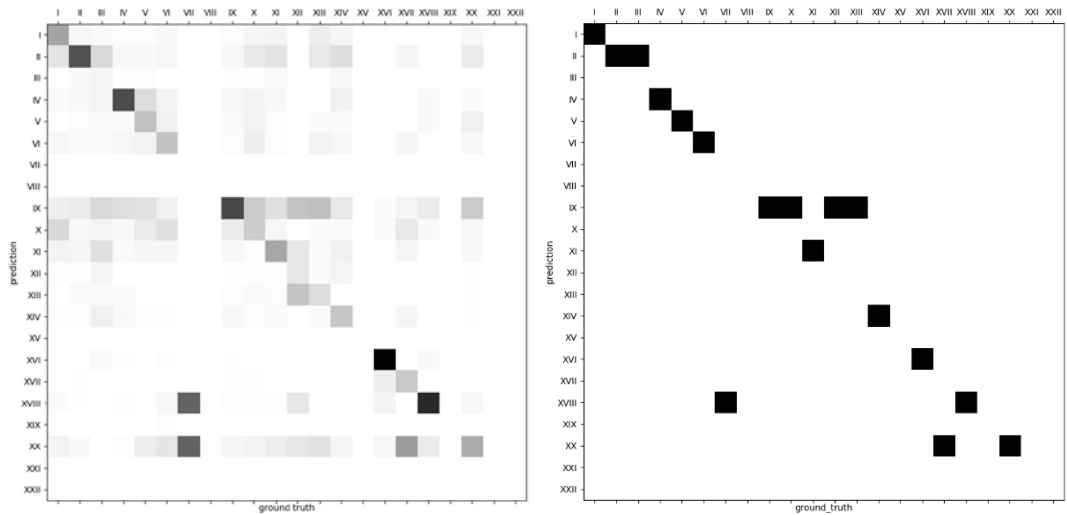

Fig. 9 Left: distribution of wrong predictions per ICD-10 chapter versus their ground truth (the lighter the rarer). Right: Same distribution's modes. Apparent missing values in both figure correspond to chapters either not represented in the test dataset or on which no mistakes were made

The model's error behaviour can also be investigated from a calibration fitness perspective. As aforementioned, some artificial neural network based models have been known to behave quite poorly in ill predicted cases, which could constitute a highly undesirable phenomenon when handling health data. The model being fit in a similar fashion to multinomial logistic regression, it not only yields a prediction, but an entire probability distribution over all possible ICD-10 codes. Assessing this distribution can offer powerful insight for such considerations. Typically, a "good" predictive model would be expected to show high confidence in cases where the prediction is correct, and a low one when mispredicting. Bar plots of said prediction confidences can be found in figure 10, and clearly show a strong tendency for the model to be more confident in its prediction in correctly predicted cases.

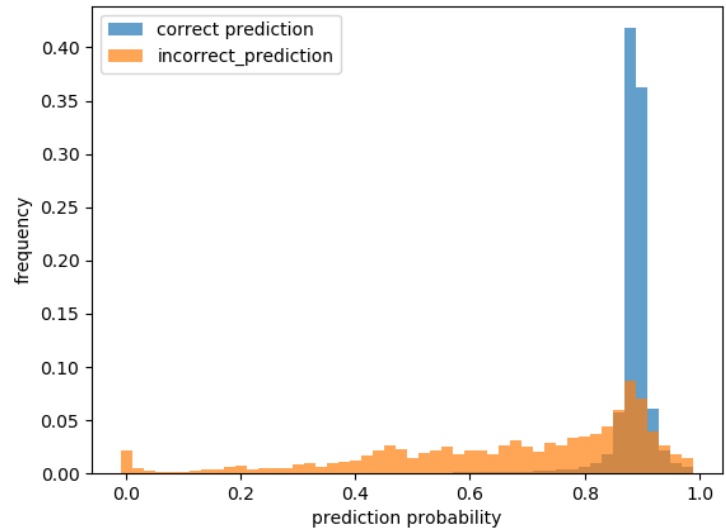

Fig. 10 histograms of prediction confidence in correct (blue) and incorrect (orange) predictions. The model typically predicts correct values with high confidence, and incorrect values with lower confidence

If predicting incorrect values with low confidence is a desirable behaviour for a predictive model, associating the ground truth with high probabilities, even in misprediction cases, should be of equal importance. This is typically assessed by evaluating whether each test set subject's corresponding ground truth is contained in the $k \in \mathbb{N}^*$ most probable values present in the model's corresponding outputted distribution. This type of metric is typically denoted as the model's top-$k$ accuracy, and helps assessing a model's ability to give high confidence to correct values, even when mispredicting. Although the academic machine learning litterature typically makes use of the top-5 accuracy in such cases, the investigated model was investigated with a top-2 accuracy only. Indeed, most death certificates present in the dataset display causal chains of events with 5 or less ICD-10 codes, with the underlying cause of death being one of them. It is consequently reasonable to expect the model to output these 5 codes as most probable, thus leading to a high but meaningless top-5 accuracy. The assessed top-2 accuracy can be found in table 2, and strongly indicates that the model consistently associate correct underlying causes of death with higher probabilities, even in ill predicted cases.

| | |
|---|---|
| Second most probable code prediction accuracy on ill predicted certificates | 0.663 [0.641, 0.685] |
| Proposed model Top-2 accuracy | 0.993 [0.992, 0.993] |

Table 2 Accuracies on codes wrongly predicted by the proposed model, and model top-2 accuracy

A richer, although more time consuming, error analysis can be derived from human observation of each error cases by an underlying cause of death coding specialist. To do so, 96 of the 1777 ill predicted death certificates in the test set were selected at random and shown to the medical practitioner referent and final decision-maker on underlying cause of death coding in France, who gave for each of the selected certificates:

- Her personal opinion of what each certificate's corresponding underlying cause should be
- A qualitative comment on the investigated model's error

The aforementioned underlying causes obtained were then confronted with both the actual values contained in the dataset and those predicted by the derived model, leading to the following observations:

- In 41% of cases, the referent agreed with the model's predictions
- In 37% of cases, the referent agreed with the underlying cause present in the dataset
- In 22% of cases, the referent agreed with neither of them

Consequently, the derived predictive model's coding can be considered as comparable in quality to the actual process responsible for the production of the investigated dataset's. In addition, a qualitative analysis of the medical practitioner's comments on the model's mistakes showed that 30% of errors committed by the predictive model are related to casuistic exceptions in coding rules, such as non-acceptable codes as underlying cause of death. Such an observation strongly reinforce the hypothesis that a hybrid, expert system deep learning approach should improve the presented system's coding accuracy.

# 5     Practical application: Recoding the 2012 French overdose anomaly

The topic of overdose related death monitoring has recently drawn attention of public health agencies around the world, specifically in light of the opioid related sanitary crisis recently witnessed in the US. Causes of death data constitutes an information source of choice to investigate such topics. In France, the CépiDc database was used to assess the evolution of overdose related deaths from 2000 to 2015, by counting for each year the number of deaths associated with the following underlying causes:

- Opioid and Cannabis related disorders (ICD-10 codes beginning with F11 and F12)
- Cocaine, hallucinogen and other stimulant related disorders (F14 to F16)
- Other psychoactive substance related disorders (F19)
- Accidental poisoning by and exposure to narcotics and psychodysleptics, not elsewhere classified (X42)
- Intentional self-poisoning by and exposure to narcotics and psychodysleptics, not elsewhere classified (X62)
- Poisoning by and exposure to narcotics and psychodysleptics, not elsewhere classified, undetermined intent (Y12)

The resulting trajectory can be found in figure 11, and shows a significant decline in overdose related deaths in 2011 and 2012. Some explanations were found for this punctual reduction, such as decrease in heroin purity[27] and heroin overdose related deaths[28], in the same time period. However,

Although this punctual reduction can be at least partially explained by observed decreases in both heroin purity[27] and heroin overdose related deaths[28] in the same time period, confrontation with results obtained from an independent source, the DRAMES dataset, suggests another hypothesis. The DRAMES study constitutes a non-exhaustive inventory of overdose related deaths detected in French Legal Medicine Institutes. As a non-exhaustive database, its death count should not exceed the value obtained from the CépiDc database. As can be seen in figure 11, this logical assertion is true for all years from 2009 to 2013 but the notable exception of 2012. This discrepancy might be explained by a coding process deficiency, a hypothesis that can easily be verified by recoding every certificate from 2012, and comparing the number of overdose related deaths in both situations.

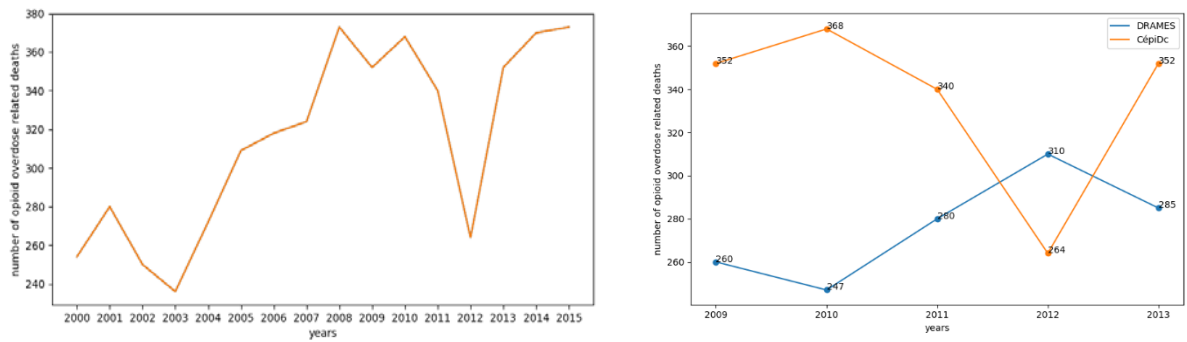

Fig. 11 Left: Evolution of overdose related deaths from 2000 to 2015 in France. The sudden decrease in 2012 appears anomalous. Right: Comparison with DRAMES data, a non-exhaustive, independent data source, which finds more deaths in 2012 than the exhaustive CépiDc database

The model derived in the previous experiment was used to recode every French death certificate from 2000 to 2015, with the year of coding set to 2015 to prevent any discrepancy related to coding rule variation. The overdose related death were then selected from the predicted underlying cause of deaths following the aforementioned methodology.

The resulting curve can be seen in figure 12, alongside the official curve, and clearly shows a smoother decrease in opioid related deaths. The discrepancy with the DRAMES database, in addition, disappears when considering the recoded underlying causes of deaths.

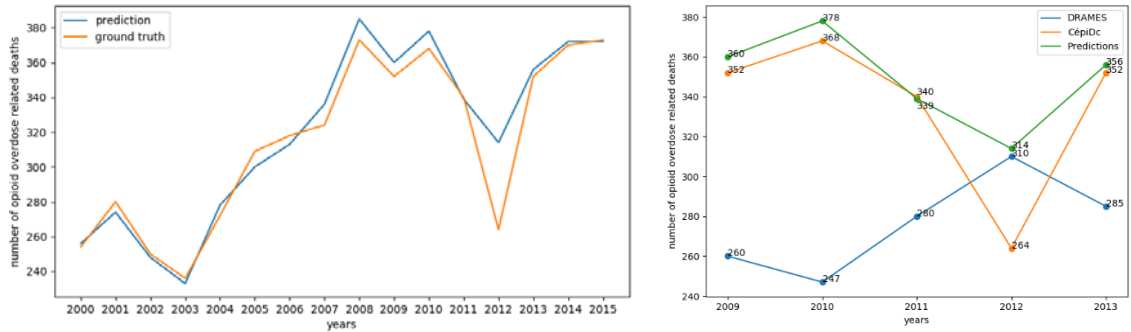

Fig. 12 Left: Evolution of opioid overdose related deaths from 2000 to 2015 in France either coded with Iris and human coders (orange) or with the proposed approach (blue). The 2012 gap, although still present, is much smoother when using predicted underlying causes. Right: Comparison with DRAMES data. The contradiction with the CépiDc database is entirely corrected with the predicted causes.

# 6 Conclusion

In this article was presented a formulation of the underlying cause of death coding from death certificate as a statistical modelling problem, who was then addressed with a deep artificial neural network, setting a new state of the art. The derived model's behaviour was thoroughly assessed following different approaches in order to identify potentially harmful biases and assess the potential of a hybrid approach mixing rule based decision system and statistical modelling. Although the proposed solution significantly outperform any other existing automated coding approaches on French death certificates, the question of model transferability to other countries requires more investigation. Indeed, the problem of distribution shift is well known in the machine learning community, and can significantly impair the model's quality[29].

The authors feel confident the model should perform with similar predictive power on other countries' death certificate with little to no supplementary effort necessary, even though this claim requires some experimental validation, unrealisable without international cooperation. To conclude, this article shows that deep artificial neural network are perfectly suited to the analysis of electronic health records, and can learn directly from voluminous dataset a complex set of medical rules, without any explicit prior knowledge. Although not entirely free from mistakes, the derived algorithm constitutes a powerful decision making tool able to handled structured, medical data with unprecedented performances, and we strongly believe that the methods developed in this article are highly reusable in a variety of settings related to epidemiology, biostatistics, and the medical sciences in general.

# A deep artificial neural network based model for underlying cause of death prediction from death certificates, Annex

## 1 Model architecture

The model architecture is mostly inspired from the Inception v2 network (1), with the following modifications:

- The Inception v2 network's first stage has been replaced with two temporal blocks (2) with successive dilation rates of 1 and 2
- Drop-out (3) , layer normalization (4) and residual connections (5) were applied to each block as can be seen on figures 1, 2 and 3
- The Inception maximum pooling operation was limited to the grid's width dimension as can be seen on figure 4
- The final softmax operation is tied to the linear embedding as described in (6)

The model's full structure is described in table 1 and figures 1 through 4.

| Type | Layer size | Dilation rate, stride or remarks | Input size |
|---|---|---|---|
| Linear embedding | 512 | _ | $6 \times 20 \times 7404$ |
| Temporal block (2) | 512 | 1 | $6 \times 20 \times 512$ |
| Temporal block | 512 | 2 | $6 \times 20 \times 512$ |
| 3 x Inception block 1 (1) | 512 | As in figure 1 | $6 \times 20 \times 512$ |
| Inception pooling (1) | 1024 | As in figure 4 | $6 \times 20 \times 512$ |
| 5 x Inception block 2 (1) | 1024 | As in figure 2 | $6 \times 9 \times 1024$ |
| Inception pooling | 1536 | As in figure 4 | $6 \times 9 \times 1024$ |
| 2 x Inception block 3 (1) | 1536 | As in figure 3 | $6 \times 4 \times 1536$ |
| Full maximum pooling | 1536 | $6 \times 4$ | $6 \times 4 \times 1536$ |
| Linear layer | 512 | _ | $1 \times 1 \times 1536$ |
| Tied linear embedding (6) | 7404 | Transpose of the first linear embedding matrix | $1 \times 1 \times 512$ |

Table 1: Model architecture and corresponding hyperparameters

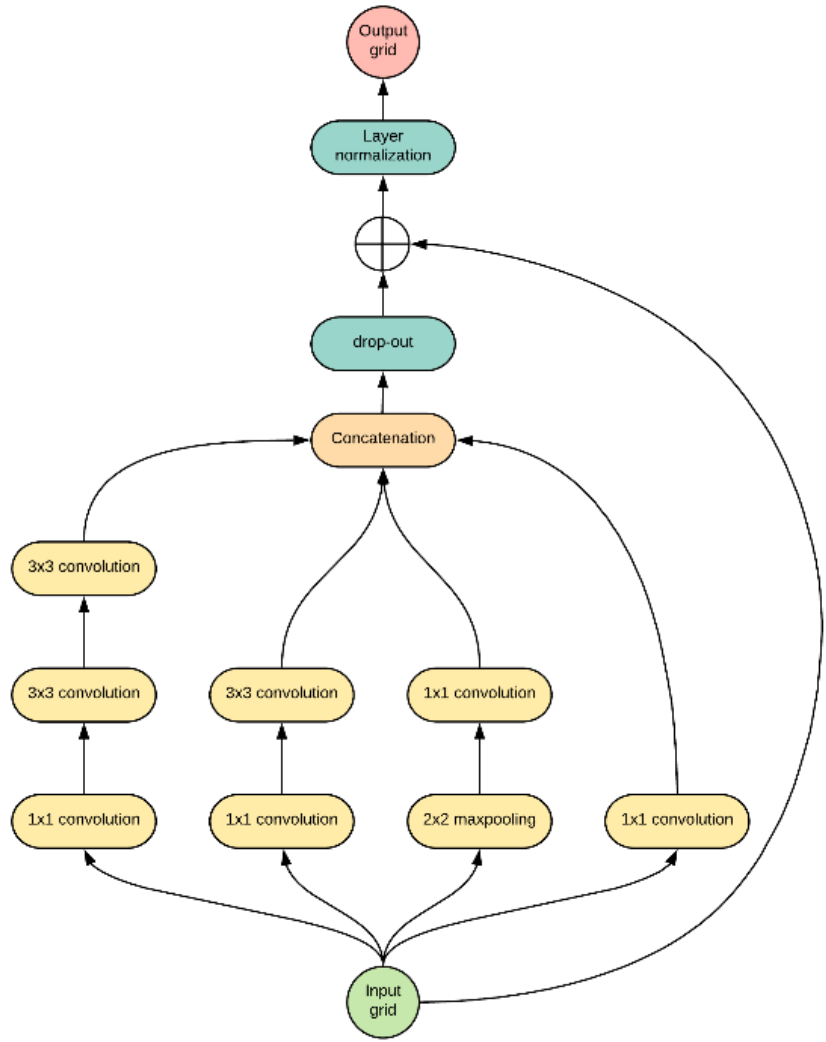

Figure 1: Inception block 1 as described in (1) with additional drop-out, layer normalization and residual connection mechanisms

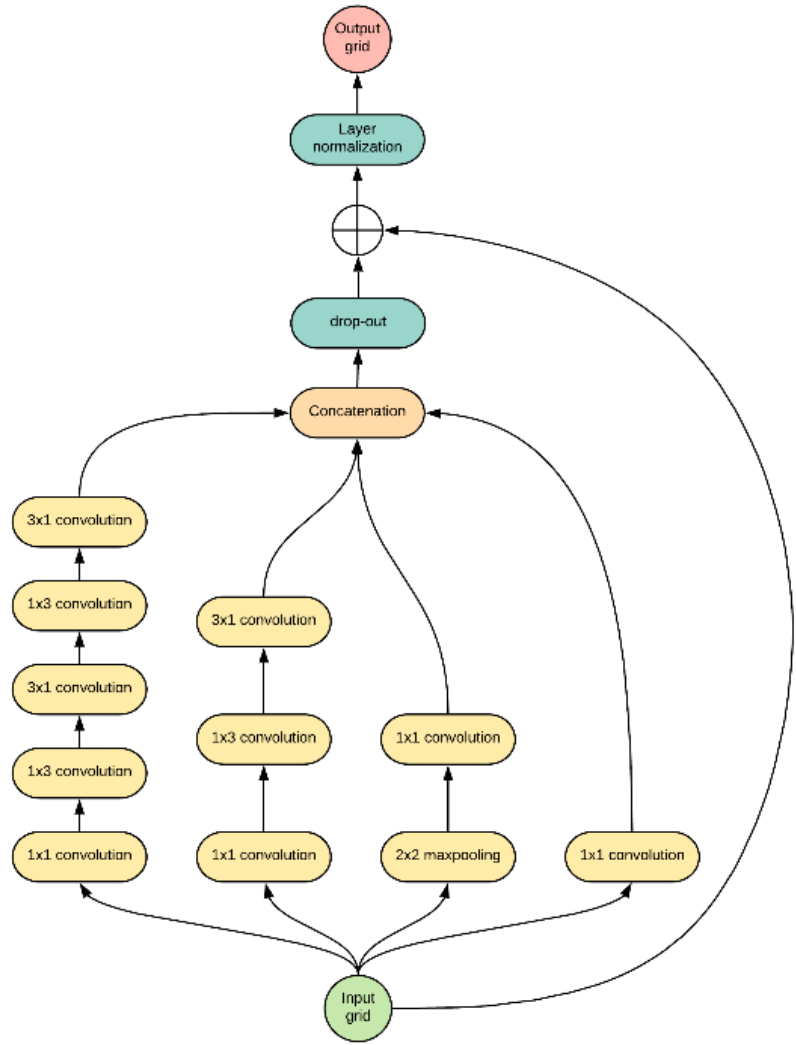

Figure 2: Inception block 2 as described in (1) with additional drop-out, layer normalization and residual connection mechanisms

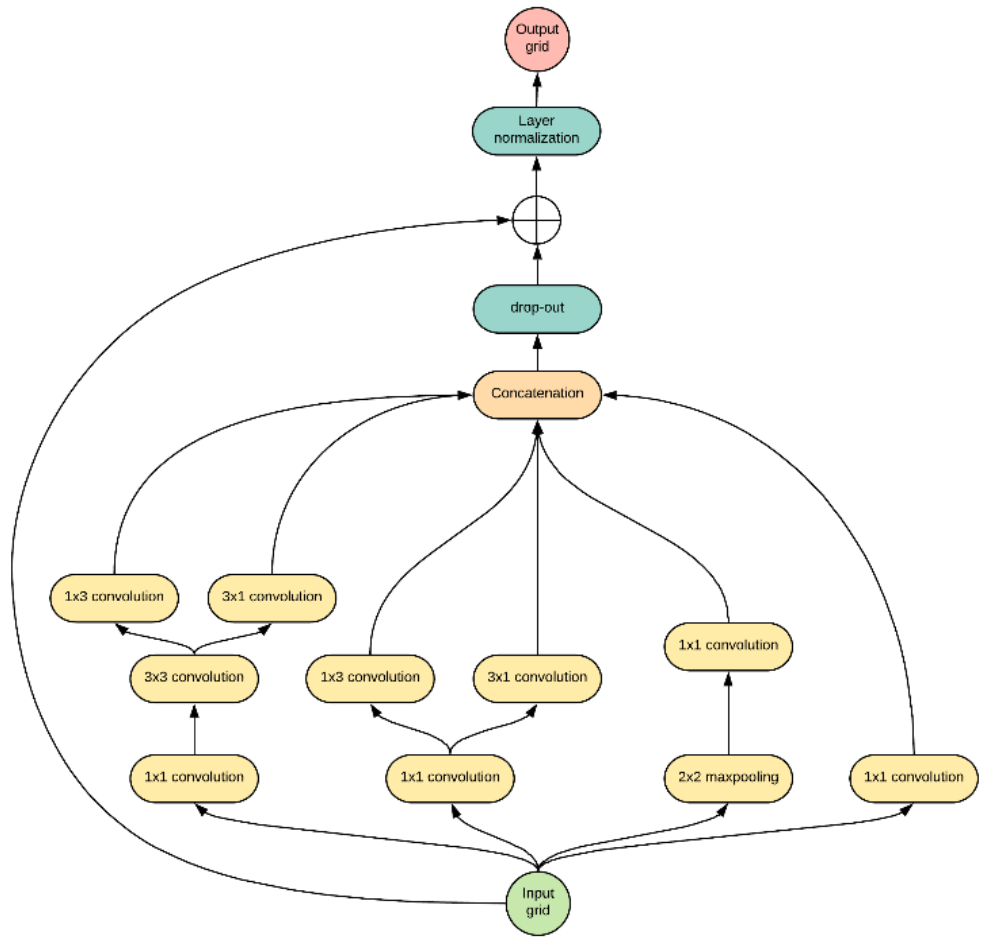

Figure 3: Inception block 3 as described in (1) with additional drop-out, layer normalization and residual connection mechanisms

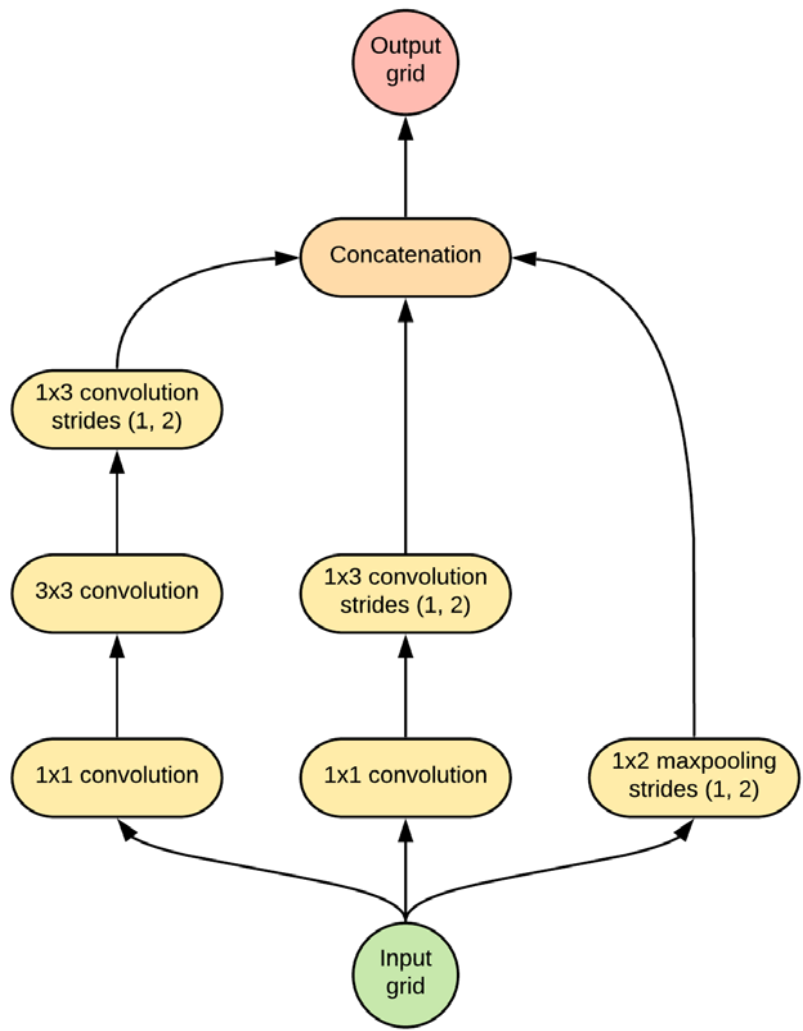

Figure 4: Inception pooling as described in (1). All grid-reducing operations are limited to the width dimension

## 2  Training methodology

The model was implemented with Tensorflow, a python-based distributed machine learning framework, on two NVidia RTX 2070 GPUs simultaneously using a mirrored distribution strategy. Training was performed using a variant of stochastic gradient descent, the Adam optimization algorithm.

The descent's step size (also called learning rate in the machine learning academic literature) was updated in real time during training according a rule defined in (7), that can be seen in figure X and is defined according to the formula:

$$Step\_size(t) = \alpha \cdot \min(t^{-0.5}, t \cdot warmup\_steps^{-1.5}) \quad \forall t \in \mathbb{R}^+$$

With:

- $\alpha \in \mathbb{R}$ a constant considered as a model hyper-parameter and defining the learning rate's overall amplitude

- $warmup\_steps \in \mathbb{N}$ another hyper-parameter defining the learning rate's linear warmup phase length

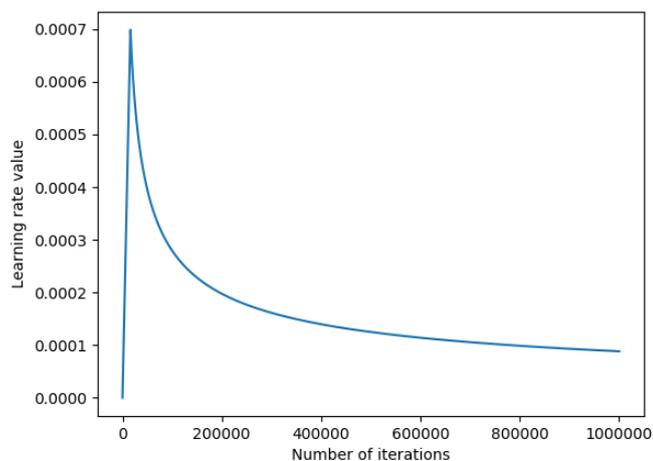

Fig. X learning rate evolution with gradient descent iterations. The learning rate follows a first linear increase warmup phase followed by an inverse root square decay

In order to limit gradient explosion phenomena typically encountered in deep neural network, the optimization was in addition controlled using gradient clipping. Essentially, the norm of all gradients computed during the descent were normalized to be of global norm equal or less than 0.1.

In addition, label smoothing was applied to the cross entropy loss to further regularize the model.

The final hyper parameters were chosen from a random search selection process with the following values:

- Batch size: 250
- Drop-out selection rate: 0.1 for all layers
- Label smoothing parameter: 0.1
- Initial learning rate constant: $\frac{2}{\sqrt{512}} \approx 0.088$
- Learning rate warmup steps: 16000
- Trainable variable initialization: Uniform variance scaling initializing